%% file: acl_latex.tex
\pdfoutput=1

\documentclass[11pt]{article}

\usepackage[preprint]{acl}

\usepackage{times}
\usepackage{latexsym}

\usepackage[T1]{fontenc}

\usepackage[utf8]{inputenc}

\usepackage{microtype}

\usepackage{inconsolata}

\usepackage{graphicx}
\usepackage{xcolor}
\usepackage{amsmath}

\usepackage{hyperref}
\usepackage{diagbox}
\usepackage{arydshln} 

\usepackage{booktabs} 
\usepackage{multirow} 
\usepackage{caption} 
\usepackage{kotex}

\title{A Method for Detecting Legal Article Competition for Korean Criminal Law Using a Case-augmented Mention Graph}

\author{Seonho An$^{1,2}$, Young Yik Rhim$^{1,3}$, Min-Soo Kim$^{1,2,\ast}$ \\
        KAIST, Republic of Korea$^1$ \quad Infolab$^2$ \quad Intellicon$^3$\\
        \texttt{\{asho1, rhims, minsoo.k\}@kaist.ac.kr}}

\begin{document}
\maketitle
\begin{abstract}
As social systems become increasingly complex, legal articles are also growing more intricate, making it progressively harder for humans to identify any potential competitions among them, particularly when drafting new laws or applying existing laws. 
Despite this challenge, no method for detecting such competitions has been proposed so far.
In this paper, we propose a new legal AI task called \textit{Legal Article Competition Detection}\,(LACD), which aims to identify competing articles within a given law.
Our novel retrieval method, CAM-Re2, outperforms existing relevant methods, reducing false positives by 20.8\% and false negatives by 8.3\%, while achieving a 98.2\% improvement in precision@5, for the LACD task.
We release our codes at \href{https://github.com/asmath472/LACD-public}{\texttt{https://github.com/asmath472/\\LACD-public}}.
\end{abstract}

\def\thefootnote{$\ast$}\footnotetext{Corresponding author.}
\renewcommand{\thefootnote}{\arabic{footnote}}

\input{latex/sec1_intro}

\input{latex/sec2_preliminaries}
\input{latex/sec3_methodology}

\input{latex/sec4_experiment-settings}

\input{latex/sec5_results}
\input{latex/sec6_relatedworks}
\input{latex/sec7_conclusions}

\input{latex/sec8_limitations}
\input{latex/sec9_ethics}

\bibliography{custom}

\appendix

\input{latex/secA_appendix}

\end{document}

%% file: latex/sec1_intro.tex
\section{Introduction}
\label{sec:intro}

In many countries, courts judge legal cases based on a law of their country, and thus, many lawyers utilize legal articles (also known as \textit{codes}, or \textit{provisions}) in their works.
In the legal AI field, several works that utilize legal articles have been suggested to solve legal AI tasks, such as Legal Judgment Prediction~\cite{feng2022legal, feng-etal-2022-legal, deng-etal-2023-syllogistic, liu2023ml}, Legal Article Retrieval~\cite{louis-spanakis-2022-statutory, paul2022lesicin, louis-etal-2023-finding}, and Legal Question Answering~\cite{Holzenberger2020ADF, louis2024interpretable}. 

\begin{figure}[t!]
    \centering
    \includegraphics[width=\columnwidth]{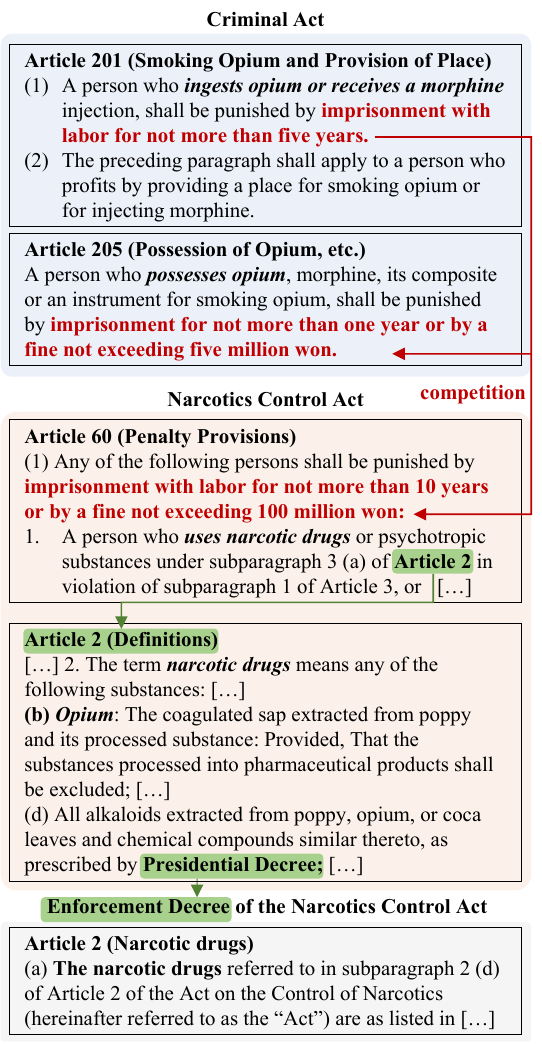}
    \caption{Example of competing legal articles in Republic of Korea, translated from Korean. In Narcotics Control Act Article 60, \textit{\textbf{uses narcotic drugs}} includes \textit{\textbf{ingests opium}}, from Article 2 of the same act. Other articles about opium are omitted.
    }
    \label{fig:example}
\end{figure}

Despite the crucial role of legal articles, some of them \textit{compete} with each other~\cite{Yoon2005Grund, kim2005Fallgruppen, araszkiewicz2021identification}.
Here, competition in legal articles refers to instances where overlapping directives or conflicting interpretations arise.
For example, in Figure~\ref{fig:example}, Article 60 of Narcotics Control Act and Article 201 of Criminal Act define different punishments for the same crime, \textit{using opium or morphine}, and therefore compete with each other.

If two articles compete with each other in a given circumstance, one is disregarded in the judgment, leading to confusion in the application of the law~\cite{Yoon2005Grund}.
Detecting such competitions is therefore essential for individuals involved in drafting laws\,(e.g.,  congress members) or applying them\,(e.g., public prosecutors).
As laws become increasingly complex\,\cite{coupette2021measuring}, manually identifying competing articles has become more challenging. 
Furthermore, with the rise of AI agents that rely on natural language rules\,\cite{bai2022constitutional, hua2024trustagent}, automating competition detection has become increasingly important.
This study aims to address this problem by developing methods to automatically detect competing articles using NLP techniques, with a particular focus on competitions within the Criminal Law of the Republic of Korea\,(hereafter referred to as \textit{Korean Law}).

We introduce a new legal AI task called \textbf{Legal Article Competition Detection\,(LACD)}, which aims to retrieve competing articles for a given article.
For example, as illustrated in Figure~\ref{fig:example}, when Criminal Act Article 201 is provided as input, the task  should identify and retrieve Criminal Act Article 205 and Narcotics Control Act Article 60.

\begin{figure*}[t!]
    \centering
    \includegraphics[width=\textwidth]{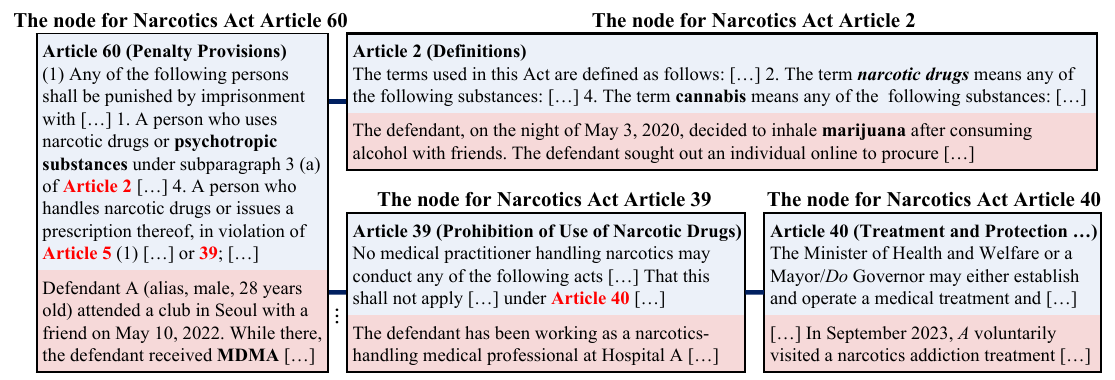}
    \caption{The example of CAMGraph. Blue and yellow boxes mean articles and corresponding LLM-generated cases, respectively. All contents are translated from Korean. 
    }
    \label{fig:camgraph-example}
\end{figure*}

For document retrieval tasks, various methods, such as TF-IDF, BM25, and DPR\,\cite{robertson2009probabilistic, karpukhin-etal-2020-dense}, have been widely used.
Recently, a \textit{retrieve-then-rerank} approach, which refines top-ranked documents from retrieval by using another language models\,(LMs), has shown high performance and low latency in retrieval tasks\,\cite{wu-etal-2020-scalable, zhu-etal-2023-learn}.
However, due to the unique characteristics of legal articles, existing retrieval methods face two significant challenges in the LACD task.

The first challenge\,(\textbf{Challenge~1}) arises from the similarity in descriptions across multiple articles, which makes it difficult for LMs to distinguish between them\,\cite{xu-etal-2020-distinguish, 10.1145/3626772.3657879}.
For example, in Figure~\ref{fig:example}, Article 201 and Article 205 of the Criminal Act share nearly identical textual descriptions, differing only in actions\,(e.g., \textit{ingests} and \textit{possesses}) and punishments\,(e.g., \textit{imprisonment} and \textit{fine}).
Consequently, an LM may infer these two articles are almost identical.
To address this challenge, some methods leverage legal cases, which provide detailed contextual descriptions, to improve differentiation\,\cite{HierSPCNet, 10.1145/3404835.3462826, paul2022lesicin}.
However, many articles lack corresponding legal cases\,(e.g., draft articles). 
This issue is referred to as the \textit{null case problem}.

The second challenge\,(\textbf{Challenge~2}) lies in the insufficiency of textual descriptions in legal articles, particularly when interpreting terminologies.
Legal articles often rely on references to other articles to define specific terms or conditions\,\cite{bommarito2010mathematical, katz2020complex}.
For example, in Figure~\ref{fig:example}, Narcotics Control Act Article 60 uses the term \textit{narcotic drugs}, which is defined in an explicitly referenced article\,(i.e., \textit{mentioned article}), Narcotics Control Act Article 2. 
Moreover, interpreting an article often requires traversing beyond direct references\,(i.e., 1-hop) to indirect references\,(n-hop). 
For example, accurately interpreting the term \textit{narcotic drugs} may require consulting the Enforcement Decree of the Narcotics Control Act.
Thus, to properly reason about legal articles, a retrieval model must leverage not only their textual description but also the explicit reference relationships\,(i.e., \textit{mention relationships}) between laws\,\cite{katz2020complex}.
However, to the best of our knowledge,
no existing studies have proposed utilizing mention relationships for legal article retrieval.

To address these two challenges in the LACD task, we propose a novel retrieve-then-rerank retriever, \textbf{CAM-Re2}, powered by the \textit{Case-Augmented Mention Graph}\,(\textbf{CAMGraph}). 
CAMGraph, introduced in this paper, is a graph-based representation of legal articles that incorporates both contextual information from cases and mention relationships.
Each legal article is represented as a node, enriched with an LLM-generated case to provide contextual depth, while edges capture mention relationships between articles, enabling reasoning across mention relationships.  
CAMGraph, built on Korean Law, comprises 192,974 nodes and 339,666 edges.
Figure~\ref{fig:camgraph-example} shows a small section of CAMGraph.
Our retriever, CAM-Re2, leverages CAMGraph to tackle the two LACD challenges effectively.
For the first challenge, CAM-Re2 replaces plain articles with CAMGraph's nodes enriched with LLM-generated cases, thereby enhancing the retriever's ability to distinguish between similar legal articles and addressing the null case problem.
For the second challenge, CAM-Re2 employs a Graph Neural Network\,(GNN) on CAMGraph, utilizing the output embedding of CAMGraph's nodes to leverage the mention relationships between articles.
This enhances the retriever's ability to interpret legal articles based on their interconnected legal contexts. 
We have constructed a dedicated dataset for training and evaluating the LACD task, which consists of 293 competing article pairs and 2,046 non-competing article pairs.
Our retriever, CAM-Re2, achieves significant improvements over the state-of-the-art retrieve-then-rerank methods, recording 20.8\% fewer false positives, 8.3\% fewer false negatives, and a 98.2\% improvement in the precision@5 score for the LACD dataset.


%% file: latex/sec2_preliminaries.tex
\section{Preliminaries}
\label{sec:preliminaries}

\subsection{Definitions}
\label{subsec:definitions}

In this section, we define the terms including \textit{case}, \textit{rule}, \textit{article}, \textit{competition}, and \textit{mention}, largely based on the definitions provided by Araszkiewicz et al.~\cite{araszkiewicz2021identification}. 
Their notations are summarized in Table \ref{tab:symbols}. 
Hereafter, we will use a \textit{legal article} and \textit{article} interchangeably.

\begin{table}[htb!]
\renewcommand{\arraystretch}{1.25}
\centering \resizebox{\columnwidth}{!}{
\begin{tabular}{c|l}
\hline
\textbf{Notation} & \textbf{Description} \\ \hline
\textit{a} & A legal article\\
\textit{c} & A case \\
\textit{p} & A judgment for cases$^\ast$\\
\textit{r} & A rule$^\ast$\\
$\mathcal{A}$ & A set of all legal articles \\
$\mathcal{X}$ & A set of proposition for cases$^\ast$\\
 \textit{compete($r_1,r_2$)}&A rule $r_1$ and $r_2$ compete with each other$^\ast$\\
 \textit{compete($a_1, a_2$)}&An article $a_1$ and $a_2$ compete with each other\\
\hline
\end{tabular}
}
\caption{Summary of notations. $^\ast$indicates a definition sourced from Araszkiewicz et al~\cite{araszkiewicz2021identification}.}
\label{tab:symbols}
\end{table}


\noindent \textbf{\textit{Definition~1\,(Case, Rule, and Article).}} 
A \textbf{case} $c$ is a sentence describing the facts of an event\,\cite{shao2020bert, sun2023law}.
A \textit{proposition}\,(denoted as $x$) for a case represents an implicit question about its facts.
A \textbf{rule} $r$ is the implicit unit of laws, consisting of a set of propositions\,(denoted as $\mathcal{X}$) and a judgment $p$ for cases $\mathcal{C}$. 
A rule $r$ judges a case $c\in \mathcal{C}$ as $p$ if and only if all propositions in $\mathcal{X}$ hold true in $c$.
We denote a rule with $\mathcal{X}$ and $p$ as $r=\textit{rule}(p, \mathcal{X})$.
An \textbf{article} $a$ is an explicit unit of laws that implicitly contains one or more rules, denoted as $r_i \sqsubseteq a$, and is expressed in sentences. 
Each article is explicitly included in one act.


\noindent \textbf{\textit{Example~1.}} We can represent Criminal Act 205\,(Article 205 in \textit{Criminal Act}) in Figure~\ref{fig:example} and its example case $c_1$ as follows:
\begin{align*}
    c_1&= \text{Bob smoked opium in his house.}\\
    a_1 &= \begin{aligned}
        &\texttt{Criminal Act Article 205 (Posses-}\\
        &\texttt{sion of Opium, ... million won.}
    \end{aligned}\\
    \mathcal{X}_1&= \left\{
    \begin{aligned}
        &\text{Is a person possesses \textit{something}?} \\
        &\text{Is \textit{something} $\in$ opium $\vee$ morphine $\cdots$?}
    \end{aligned}\right\}\\
    p_1&=\text{Less than five million won fine} \\
        &\vee \text{Less than one year imprisonment.} \\ 
    r_1&=\textit{rule}(p_1, \mathcal{X}_1)\textit{, }r_1\sqsubseteq a_1
\end{align*}
Since Bob \textit{smoked}\,(a proposition about possession in $\mathcal{X}_1$) \textit{opium}\,(a proposition regarding $\text{opium} \vee \text{morphine} \vee \cdots$ in $\mathcal{X}_1$), all propositions in $\mathcal{X}_1$ hold in $c_1$, and thus, case $c_1$ is judged as $p_1$. 

\vspace*{0.2cm}

\noindent \textbf{\textit{Definition~2\,(Competition).}}
\begin{enumerate}
    \item Two rules compete, i.e., $\textit{compete}$ $(\textit{rule}(p_1,$ $\mathcal{X}_1), \textit{rule}(p_2, \mathcal{X}_2))$ if and only if
    $p_1\ne p_2$, and $\mathcal{X}_1$ includes $\mathcal{X}_2$, or vice versa.
    
    \item If two rules compete, then the articles containing those rules also compete.
    Specifically, \textit{compete}$\text{(}a_1, a_2\text{)}$ if $\textit{compete}\text{(}r_i,r_j\text{)}$, $r_i \sqsubseteq a_1$, and $r_j \sqsubseteq a_2$.
\end{enumerate}

\noindent \textbf{\textit{Example~2.}} We can represent Criminal Act 201 in Figure~\ref{fig:example} as follows:

\begin{align*}
    a_2 &= \begin{aligned}
        &\texttt{Criminal Act Article 201 (Smoking} \\
        &\texttt{Opium and Provision ... morphine.}
    \end{aligned}\\
    \mathcal{X}_2&=\left\{
    \begin{aligned}
        &\text{Is a person uses \textit{something}?} \\
        &\text{Is \textit{something} $\in$ opium $\vee$ morphine?}
    \end{aligned}\right\}\\
    p_2&=\text{Labor not more than five years.} \\ 
    r_2 &= \textit{rule}(p_2, \mathcal{X}_2)\textit{, }r_2 \sqsubseteq a_2
\end{align*}
Here, \textit{possesses} $\mathcal{X}_1$ includes \textit{uses} $\mathcal{X}_2$, and \textit{opium}$\vee$ \textit{morphine} $\cdots$ $\mathcal{X}_1$ includes \textit{opium} $\vee$ \textit{morphine} $\mathcal{X}_2$, establishing that $\mathcal{X}_1$ includes $\mathcal{X}_2$. 
Since $p_1 \ne p_2$, rules $r_1$ and $r_2$ compete, and consequently, articles $a_1$ and $a_2$ also compete.

\vspace*{0.2cm}

\noindent \textit{\textbf{Definition~3\,(Mention).}}
If an article $a_1$ explicitly cites another article $a_2$, then $a_1$ \textit{mentions} $a_2$. 

\noindent \textit{\textbf{Example~3.}}
In Figure~\ref{fig:example}, Narcotics Control Act Article 60 mentions Article 2 of the same act.

\subsection{Retrieve-then-rerank methods}
\label{subsec:three-retrieval-approaches}


Given a query article $a_q$, retrieve-then-rerank methods\,\cite{nogueira2019passage, wu-etal-2020-scalable, glass-etal-2022-re2g, zhu-etal-2023-learn, song-etal-2024-re3val} retrieve a set of articles $\mathcal{A}_{\textit{ret}}$ through the following three steps.
\begin{align*}
    \text{\textbf{Step 1.} }&\mathbf{v}_{a_q} = \textit{enc-bi}(a_q), \mathbf{v}_{a}=\textit{enc-bi}(a)(a\!\in\! \mathcal{A})\\
    \text{\textbf{Step 2.} }&\mathcal{A}_{\textit{topk}} = \{a\mid \text{top-k} \text{ by } \text{sim}(\mathbf{v}_{a_q}, \mathbf{v}_{a})\}\\
    \text{\textbf{Step 3.} }&\mathcal{A}_{\textit{ret}}\!=\!\{a_i\mid p(a_q,a_i)\!>\!\theta\}\textit{ }(a_i\!\in\! \mathcal{A}_{\textit{topk}})\\
    \textit{where }& p(a_q,a_i) = \textit{probcalc}(\textit{enc-cross}(a_q \oplus a_i))
\end{align*}
Here, $\mathbf{v}_a$ is the vector representation of article $a$; \textit{sim} presents a similarity function, such as cosine; \textit{probcalc} refers to a layer for calculating retrieval probability; $\oplus$ denotes a textual concatenation operator. 
We denote the retrieval probability of article $a_i$ for the query $a_q$ as $p(a_q, a_i)$.


In Step 1, each article $a \in \mathcal{A}$ is pre-encoded into vector representations $\mathbf{v}_{a}$ using a bi-encoder. The bi-encoder also encodes the query article $a_q$ into a vector representation $\mathbf{v}_{a_q}$.  
In Step 2, the retriever selects the top-k articles $\mathcal{A}_{\textit{topk}}$ based on the similarity function $\text{sim}(\mathbf{v}_{a_q}, \mathbf{v}_{a})$. 
In Step 3, the cross-encoder processes each article $a_i \in \mathcal{A}_{\textit{topk}}$ together with \(a_q\), and inputs it into the \textit{probcalc} layer to compute the retrieval probability $p(a_q,a_i)$.
The retriever returns the set of articles $\mathcal{A}_{\textit{ret}}$, consisting of those articles $a_i$ for which $p(a_q, a_i)$ exceeds a threshold~$\theta$.  



%% file: latex/sec3_methodology.tex
\section{Methodology}

\label{sec:method}
\subsection{The LACD task}
We define Legal Article Competition Detection\,(LACD) as a retrieval task that takes a query article $a_q$ as input, and retrieve a set of competing articles, $\{a_i|\textit{compete}(a_q,a_i)\}$, as output.
This definition is particularly useful for identifying articles that compete with one currently being drafted.
A naïve method to identify all competing articles within a law involves performing the LACD task for each article in the law.

We will refer to the retrieve-then-rerank method described in Section~\ref{subsec:three-retrieval-approaches} as the \textit{naïve Re2} retriever.
Unlike traditional retrieval tasks such as open-domain QA\,\cite{karpukhin-etal-2020-dense}, which primarily focus on identifying related documents,
the LACD task, as discussed in Section~1, presents unique challenges. 
Consequently, the naïve Re2 retriever often retrieves textually similar but semantically irrelevant and non-competing articles.

\begin{figure*}[tb!]
    \centering
    \includegraphics[width=\textwidth]{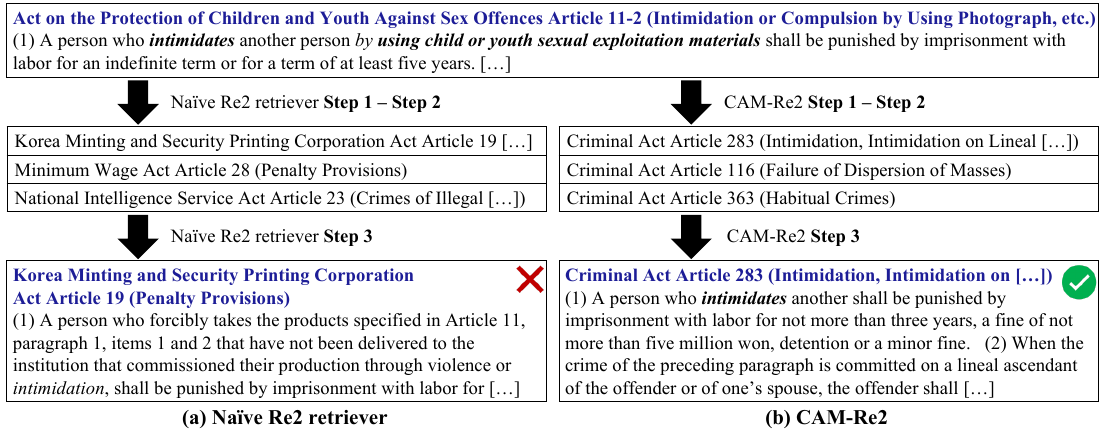}
    \caption{LACD results of (a) the naïve Re2 retriever and (b) our CAM-Re2 retriever. The query, Act on the Protection Of Children and Youth Against Sex Offenses Article 11-2, is \textbf{a draft article} in September 30, 2024\,(i.e., not yet legalized). Both retrievers are implemented using the KoBigBird model~\cite{jangwon_park_2021_5654154}, and $k=3$ is used for top-k selection.
    }
    \label{fig:newlaw}
\end{figure*}

For example, Figure~\ref{fig:newlaw}(a) shows the challenges faced by the naïve Re2 retriever in the LACD task.
We detail the issues encountered at each retrieval step for the query, Act on the Protection Of Children and Youth Against Sex Offenses Article 11-2.
At Step~1, articles that are textually similar but semantically unrelated to $a_q$\,(e.g., Korea Minting and Security Printing Corporation Act Article 19) are mapped to similar vectors.
Conversely, articles that are textually different but semantically related to $a_q$\,(e.g., Criminal Act Article 283) are mapped to distant vectors. 
As a result, at Step~2, the retriever selects almost irrelevant articles as the top-k candidates $\mathcal{A}_{\textit{topk}}$ for the query article, which illustrates Challenge~1.
At Step~3, the retriever attempts to calculate the probabilities that $a_q$ competes with each $a_i \in \mathcal{A}_{\textit{topk}}$. 
This step requires a precise understanding of the concepts outlined in each article\,(e.g., `Article 11, paragraph 1, items 1 and 2') and the ability to reason about implicit inclusion relationships between the rules within the articles.
However, the retriever often fails to accurately interpret these concepts, particularly when they depend on definitions provided in other articles. 
Consequently, the retriever fails to filter out incorrect results, such as Korea Minting and Security Printing Corporation Act Article 19, thus exhibiting Challenge~2.


\subsection{CAMGraph}
\label{subsec:camgraph}


CAMGraph $\mathcal{G}$ is composed of nodes $\mathcal{N}$ and edges $\mathcal{E}$.
Detailed explanations of these components are provided below.


\noindent \textbf{Nodes:}
Each node $n \in \mathcal{N}$ can be represented as a pair $\langle a, c \rangle$, where $a$ is an article and $c$ is an \textit{associated case} in which the article $a$ can be applied. 
Articles are generally classified into two types: those with cases judged by courts\,(\textit{real case articles}) and those without\,(\textit{null case articles}). 
For null case articles, we generate a case using a prompted-LLM, $LM_\textit{case}$, and assign it as the associated case.
For real case articles, we have two options: (1) randomly select one of the real cases as the associated case, or (2) generate and assign a case as the associated case, similar to the approach for null case articles.
We observed that the second option yields better performance, since real cases often contain irrelevant information about articles and have a slightly different linguistic style compared to the cases generated for null case articles.
Therefore, we adopt the second option for node construction. 
Further details are provided in Appendix~\ref{appendix:real-case-for-camgraph}.
Formally, the nodes are expressed as follows:
\begin{align*}
    \mathcal{N}=\{\langle a_i, c_i \rangle &| a_i \in \mathcal{A},c_i=LM_{\textit{case}}(a_i)\}
\end{align*}

\noindent \textbf{Edges:}
The edges are constructed by crawling the mention relationships in Korean Law, as documented by the Ministry of Government Legislation.
All expressions of mention relationships follow specific templates~\cite{moleg2023standards}, such as `\texttt{제\textit{num}조}' (meaning `Article \textit{num}').
Based on these templates, we can construct edges. 
Copyright-related considerations regarding this process are discussed in Section~\ref{sec:ethical-considerations}.
Formally, the edges are expressed as follows:
\begin{align*}
\mathcal{E} = \{(n_i, n_j)\mid n_i=\langle a_i,c_i \rangle, n_j=\langle a_j, c_j \rangle, \\
a_i \text{ mentions } a_j \text{ or }  a_j \text{ mentions } a_i \}
\end{align*}

\noindent \textbf{Statistics:}
We build 192,974 nodes and 339,666 edges for Korean Law as of September 30, 2024. 
Detailed statistics are in Table~\ref{tab:CAMGraph-stats}.

\begin{table}[tb!]
\renewcommand{\arraystretch}{1.25}
\centering
\small
\resizebox{\columnwidth}{!}
{
\begin{tabular}{lcc}
\hline
\textbf{Statistics}&\textbf{Avg. value} & \textbf{Std. deviation}\\
\hline
\textbf{Nodes} \\
Article words count & 75.5 & 90.6 \\
Case words count & 114.0 & 29.3 \\
\hline
\textbf{Edges} & \\
Edges for nodes & 4.57 & 11.11 \\
\hline
\end{tabular}
}
\caption{Statistics about CAMGraph for Korean Law.}
\label{tab:CAMGraph-stats}
\end{table}

\subsection{CAM-Re2 retriever}
\label{subsec:our-lacd-retriever}

\begin{figure}[tb!]
    \centering
    \includegraphics[width=\columnwidth]{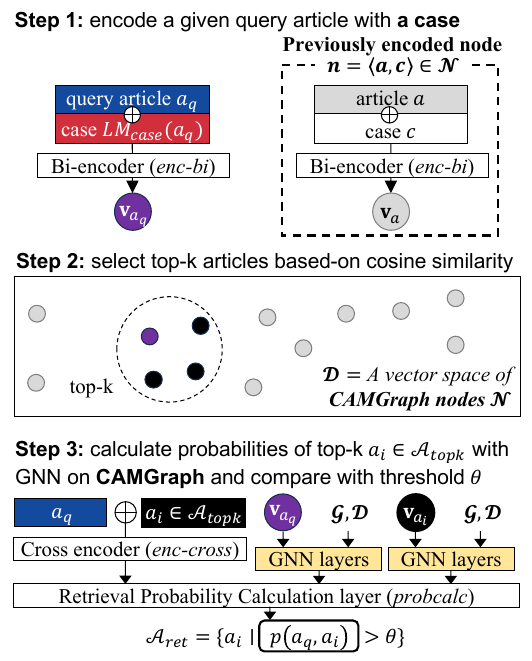}
    \caption{Overview of CAM-Re2\,(purple is the query vector).
    }
    \label{fig:camre2}
\end{figure}

The proposed CAM-Re2 retriever operates as follows: (1) it encodes nodes of CAMGraph instead of articles in Step~1, (2) selects the top-k nodes in Step~2, and (3) applies GNNs on CAMGraph in conjunction with the cross encoder in Step~3.
Thus, Steps 1 and 3 are quite different from those of the naïve Re2 retriever.
Figure~\ref{fig:camre2} shows these steps.
Formally, CAM-Re2 is described as follows:
\begin{align*}
    & \text{\textbf{Step 1.} }\!
    \begin{aligned}
        &\mathbf{v}_{a_q} = \textit{enc-bi}(a_q \textcolor{blue}{\oplus LM_{\textit{case}}(a_q)})\\
        &\mathbf{v}_{a}=\textit{enc-bi}(a\textcolor{blue}{\oplus c})\texttt{  }\textcolor{blue}{(\langle a,c \rangle \in \mathcal{N})}
    \end{aligned}\\
    &\text{\textbf{Step 2.} }\!\mathcal{A}_{\textit{topk}} = \{a \mid \text{top } k \text{ by } \text{sim}(\mathbf{v}_{a_q}, \mathbf{v}_{a})\}\\
    &\text{\textbf{Step 3.} }\!\mathcal{A}_{\textit{ret}}=\{a_i|\textit{ProbCal}(\textit{enc-cross}(a_q \oplus a_i);\\
    &\textcolor{blue}{\textit{GNN}(\mathbf{v}_{a_q}\!,\mathcal{G}\!, \mathcal{D});\textit{GNN}(\mathbf{v}_{a_i}\!,\mathcal{G}\!, \mathcal{D})})\!>\!\theta\}(a_i\!\in\!\mathcal{A}_{\textit{topk}})
\end{align*}

Here, $\mathcal{D}$ represents the input embedding vectors of the nodes of CAMGraph, and the differences from naïve Re2 are highlighted in blue.

For example, Figure~\ref{fig:newlaw}(b) shows how CAM-Re2 retrieves competing article differently from naïve Re2 across Steps 1, 2, and 3.
At Step~1, $\mathbf{v}_{a_q}$ is obtained by augmenting a query article $a_q$ with a generated case that describes relevant situations, and then bi-encoding it. 
Consequently, $\mathbf{v}_{a_q}$ is positioned closer to the vectors of semantically related nodes and farther from unrelated ones in the vector space of CAMGraph.
At Step~2, more relevant articles such as Criminal Act Articles 283, 116, and 363, are selected as the top-k nodes for the query.
At Step~3, GNN on CAMGraph enhances the semantic representation of $\{\mathbf{v}_a\}$ by aggregating critical concepts essential for reasoning and connected via mention relationships. 
As a result, CAM-Re2 effectively filter out articles that do not compete with $a_q$, such as Criminal Act Article 363, and successfully identifies the competing article with the query, Ciriminal Act Article 283.



%% file: latex/sec4_experiment-settings.tex
\section{Experimental settings}
\label{sec:experiments}

\subsection{The LACD dataset}
\label{subsec:dataset}

To build the LACD dataset for Korean Law, we collected 2,339 pairs of articles $\langle a_1, a_2 \rangle$ and manually labeled them as either competing or non-competing. 
We collected the pairs based on two criteria: 
\begin{enumerate}
    \item The article $a_1$ is in \textit{Criminal Act} and has a mention relationship with $a_2$, or vice versa.
    \item $a_1$ and $a_2$ are in one of the \textit{acts about crimes}.
\end{enumerate}
The term \textit{acts about crimes} refers to a set of acts included in the Korean Bar Exam, listed in Appendix \ref{appendix:criminal-law-acts}.
A total of 1,167 pairs were collected by the first criterion, while the remaining 1,172 pairs were gathered by the second criterion.
Detailed statistics are shown in Tables~\ref{tab:dataset-stats} and \ref{tab:article-stats}. 

\noindent \textbf{Quality review:} Our dataset was validated by legal experts. As a result, nearly 94\% of the pairs aligning with real-world competitions (see Appendix \ref{appendix:competitions-in-the-real-world}) and the remaining 6\% differing but still fitting our definitions.

\begin{table}[tb!]
\renewcommand{\arraystretch}{1.1}
\centering 
\small
\resizebox{\columnwidth}{!}
{
\begin{tabular}{lccc}
\hline
\textbf{Datasets}&\textbf{positives} & \textbf{negatives} & \textbf{Avg. words}\\
\hline
Train & 180 & 1,219 & 131.4 \\
Validation & 50 & 421 & 128.74 \\
Test & 63 & 406 & 125.03 \\
\hline
Total & 293 & 2,046 & 129.60 \\
\hline
\end{tabular}
}
\caption{Statistics for $\langle a_1, a_2 \rangle$ pairs in the dataset. 
}
\label{tab:dataset-stats}
\end{table}


\begin{table}[htb!]
\centering 
\resizebox{\columnwidth}{!}
{
\begin{tabular}{lc}
\hline
\textbf{The name of Acts} & \textbf{No. of articles \textbf{(proportion)}} \\
\hline
Criminal Act & 440 (28.3\%) \\
Youth Sexual Offenses Act & 80 (5.1\%)\\
Sexual Crimes Act & 71 (4.6\%)\\
Specific Crimes Act & 31 (2.0\%)\\
Others & 935 (60.1\%) \\
\hline
Total &  1557 (100\%)\\
\hline
\end{tabular}
}
\caption{Statistics of acts in the LACD dataset. \textit{Others} comprises 700 acts, each under 1\% proportion.}
\label{tab:article-stats}
\end{table}

\subsection{Implementation of retrievers}
\label{subsec:implementation}
We employ {KoBigBird}\,\cite{jangwon_park_2021_5654154} model as a bi-encoder; and {KoBigBird}, {Qwen2.0} \cite{yang2024qwen2}, Llama 3.2 \cite{dubey2024llama} as cross encoders.
For CAM-Re2, we utilize gpt-4o-mini\,(OpenAI) with 0.7 temperature as a case generator, and primarily employ a two-layer GATv2\,\cite{DBLP:conf/iclr/Brody0Y22} as the GNN architecture.
We compare other GNN architectures in Section~\ref{subsec:gnn-architecture-compare}.
For the vector database, we utilize Chroma DB\footnote{\href{https://www.trychroma.com/}{chroma DB official site}}.

\subsection{Training and testing}
\label{subsec:training}
We build naïve Re2 and CAM-Re2 retrievers by finetuning the bi-encoder first and the cross encoder afterward.
For each pair of articles $s = \langle a_i, a_j \rangle$ in the LACD dataset $\mathcal{S}$, the training objective $\hat{y_{s}}$ is defined as follows:
\begin{align*}
    &\text{\textbf{enc-bi.} }\hat{y_{s}}=\text{sim}(\mathbf{v}_{a_i},\mathbf{v}_{a_j}) = \frac{\mathbf{v}_{a_i} \cdot \mathbf{v}_{a_j}}{\|\mathbf{v}_{a_i}\| \|\mathbf{v}_{a_j}\|}\\
    &\text{\textbf{enc-cross.} } \hat{y_{s}}=p(a_i,a_j)=\textit{probcalc}(\cdots)
\end{align*}

The number of pairs $N=\|\mathcal{S}\|$ and the ground-truth label $y_{s}$ for each pair $s$, we utilize \textit{Binary Cross Entropy loss} as a loss function $\mathcal{L}$.
\begin{equation*}
    \mathcal{L}=-\frac{1}{N}\sum_{s \in \mathcal{S}}y_{s}\cdot \log(\hat{y_{s}})+(1-y_{s})\cdot \log(1-\hat{y_{s}})
\end{equation*}

During training, we evaluate a model approximately every 1/13th of the training steps using the validation set, and select the encoder that achieves the highest ROC-AUC score. 
Detailed parameters are in Appendix \ref{appendix:parameters}.
We also train the CAM-Re2 enc-bi on \textit{two} generated cases; detailed evaluations of utilizing multiple cases are provided in Appendix \ref{appendix:multiple-cases}.
To validate the effectiveness of CAM-Re2, we conduct two different experiments: 
\begin{enumerate}
    \item \textbf{Experiment for Step~3}: This evaluates only Step~3 component of each retriever.
    \item \textbf{Experiment for all Steps}: This evaluates the complete process of each retriever.
\end{enumerate}

As metrics, we utilize the F1 score and accuracy for the former experiment and the precision@5 score for the latter experiment.
ROC-AUC is not used as a metric, as it is employed during training for early stopping.
All experiments are conducted on a single machine equipped with eight Nvidia A100 GPUs.
For simplicity, we assume that all articles for retrieval are in Acts\,(not in Enforcement degree). Details are in Appendix \ref{appendix:law-hierarchy}.


%% file: latex/sec5_results.tex
\section{Results and analysis}
\label{sec:results}

\subsection{Experiment for Step 3}
\label{subsec:step3-results}


Since Naïve Re2 and CAM-Re2 differ at Steps 1 and 3, there are four possible combinations: (1) Naïve Re2 at all Steps\,(\textbf{Naïve Re2}), (2) Naïve Re2 at Step~1 + CAM-Re2 at Step~3\,(\textbf{N1+C3}), (3) CAM-Re2 at Step~1 + Naïve Re2 at Step~3\,(\textbf{C1+N3}), and (4) CAM-Re2 at all Steps\,(\textbf{CAM-Re2}).
Table~\ref{tab:nli-finetune-results} shows their performance results. 
CAM-Re2 achieved F1 score improvements of 9.6\%p, 6.6\%p and 3.7\%p with KoBigBird, Qwen2.0 and Llama 3.2, respectively, compared to Naïve Re2. 
Since the combination (3) does not utilize $\mathcal{D}$, its results are identical to the combination (1).
The results demonstrate the effectiveness of CAM-Re2 for the LACD task. 

\begin{table}[htb!]
\renewcommand{\arraystretch}{1}
\centering 
\small
\resizebox{\columnwidth}{!}
{
\begin{tabular}{l|cc|cc}
\hline
\multirow{2}{*}{\rule{0pt}{2.5ex}\diagbox{\textbf{Step 3}}{\textbf{Step 1}}} & \multicolumn{2}{c|}{Naïve Re2} & \multicolumn{2}{c}{CAM-Re2 \textbf{(ours)}} \\
& F1 & Acc. & F1 & Acc. \\
\hline
\textbf{KoBigBird (114M)} & & & &\\
Naïve Re2 & 48.9 & 88.0 & 48.9 & 88.0 \\
CAM-Re2 \textbf{(ours)} & 56.7 & 89.2 & \textbf{58.5} & \textbf{89.4} \\
\hdashline
\textbf{Qwen2.0 (0.5B)} & & & &\\
Naïve Re2 & 53.1 & 88.8 &  53.1 & 88.8 \\
CAM-Re2 \textbf{(ours)} & 56.7 & 87.8 & \textbf{59.7} & \textbf{90.3} \\
\hdashline
\textbf{Llama 3.2 (1B)} & & & &\\
Naïve Re2 & 58.0 & 88.2 & 58.0 & 88.2 \\
CAM-Re2 \textbf{(ours)} & 53.9 & 87.5 & \textbf{61.7} & \textbf{89.3} \\
\hline
\end{tabular}
}
\caption{Performance~(\%) of Step~3 in the LACD dataset. All experiments were conducted three times, with each reported value being an average.}
\label{tab:nli-finetune-results}
\end{table}

The N1+C3 combination is less effective than CAM-Re2, and even performs worse than Naïve Re2 when using Llama 3.2.
It is because the node encoding of CAM-Re2 significantly  improves the quality of the vector representation $\{\mathbf{v}_a\}$ compared to the article encoding of Naïve Re2. 
Enhanced vector quality is crucial because GNN in Step~3 takes these vector representations as input. 
Consequently, N1+C3 tends to yield suboptimal performance, as the initial input quality directly impacts the GNN's effectiveness. 


\subsection{Experiment for all Steps}
\label{subsec:lacd-retrieval-results}

\begin{figure}[htb!]
    \centering
    \includegraphics[width=\columnwidth]{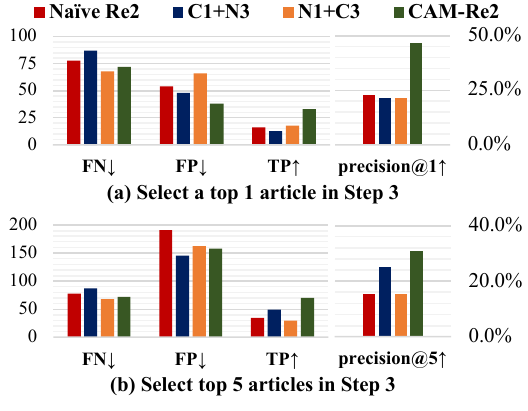}
    \caption{Performance across all Steps\,(Qwen2.0 is used as the cross encoder, and $k=10$ is used for the top-k selection).
    }
    \label{fig:retrieval-results}
\end{figure}

Figure~\ref{fig:retrieval-results} shows the performance of the four combinations for the entire process.
Specifically, Figures~\ref{fig:retrieval-results}(a) and (b) show the results when selecting the top-1 and top-5 articles, respectively.
In Figure~\ref{fig:retrieval-results}(a), CAM-Re2 reduces false negatives\,\textbf{(FN) by 7.7\%}, false positives\,\textbf{(FP) by 29.6\%}, while in Figure~\ref{fig:retrieval-results}(b), it reduces \textbf{FP by 17.28\%}.
In addition, CAM-Re2 achieves a \textbf{103\% increase in precision@1} and a \textbf{98.2\% increase in precision@5} compared to Naïve Re2.

This reduction in FP and FN of CAM-Re2 primarily stems from the improved bi-encoder performance using node encoding in Step~1.
In contrast, the bi-encoder performance of Naïve Re2 is relatively poor, leading to more FPs and misses TPs in Step~3. 
When comparing Naïve Re2\,(in red) with N1+C3\,(in orange), N1+C3 does not outperform Naïve Re2 despite utilizing GNN, due to poor encoding.
This observation aligns with the analysis in Section~\ref{subsec:step3-results}.
C1+N3\,(in blue) more frequently achieves better performance than Naïve Re2, particularly in precision@5.
However, it shows worse performance than CAM-Re2 due to the issue of insufficient textual descriptions.


\subsection{Ablation study 1: Step~3 w/o cross encoder}
\label{subsec:gnn-only-result}

In this section, we evalute the effectiveness of using GNNs at Step~3.
Table~\ref{tab:gnn-only-step3} shows the performance of Step~3 without the cross encoder, allowing for a straightforward assessment of the GNN layers' impact. 
When using the bi-encoder of Naïve Re2 at Step~1, using GNN at Step~3 improves the F1 score by 14.4\%p\,(36.1\% to 50.5\%).
When utilizing the bi-encoder of CAM-Re2 at Step~1, the use of GNN at Step~3 yields an even greater improvement in the F1 score by 17.5\%p\,(46.7\% to 64.2\%).


\begin{table}[htb!]
\renewcommand{\arraystretch}{1}
\centering
\resizebox{\columnwidth}{!}
{
\begin{tabular}{l|cc|cc}
\hline
\multirow{2}{*}{\rule{0pt}{2.5ex}\diagbox{\textbf{Step 3 Method}}{\textbf{Step 1 Method}}}  & \multicolumn{2}{c|}{Naïve Re2} & \multicolumn{2}{c}{CAM-Re2} \\
 & F1 & Acc. & F1 & Acc. \\
\hline
CAM-Re2 w/o GNN layers & 36.1 & 89.4 & 46.7 & 88.5 \\
CAM-Re2 w GNN layers & 50.5 & 89.4 & \textbf{64.2} & \textbf{90.4} \\
\hline
\end{tabular}
}
\caption{Performance of Step~3 without \textit{enc-cross}.
}
\label{tab:gnn-only-step3}
\end{table}

\subsection{Ablation study 2: GNN architectures}
\label{subsec:gnn-architecture-compare}

\begin{table}[htb!]
\centering
\renewcommand{\arraystretch}{1}
\resizebox{\columnwidth}{!}
{
\begin{tabular}{l|cc|cc}
\hline
\multirow{2}{*}{\rule{0pt}{2.5ex}\diagbox{\textbf{Step 3 Method}}{\textbf{Step 1 Method}}}  & \multicolumn{2}{c|}{Naïve Re2} & \multicolumn{2}{c}{CAM-Re2} \\
 & F1 & Acc. & F1 & Acc. \\
\hline
Naïve Re2 & 48.9 & 88.0 & 48.9 & 88.0 \\
CAM-Re2 (GCN) & 54.6 & 87.8 & 58.0 & 88.8 \\
CAM-Re2 (GraphSAGE) & 47.8 & \textbf{89.5} & 52.5 & 88.8 \\
CAM-Re2 (GATv2, \textbf{ours}) & \underline{56.7} & 89.2 & \textbf{58.5} & \underline{89.4} \\
\hline
\end{tabular}
}
\caption{Performance using different GNN architectures. 
}
\label{tab:gnn-comparison}
\end{table}

We evaluate two alternative GNN architectures for CAM-Re2: GCN~\cite{DBLP:conf/iclr/KipfW17} and GraphSAGE~\cite{DBLP:conf/nips/HamiltonYL17}.
Table~\ref{tab:gnn-comparison} shows the performance results.
Both GCN and GraphSAGE show significant improvements over Naïve Re2, emphasizing the effectiveness of GNNs in the CAM-Re2 framework. 
However, CAM-Re2 with GATv2 overall achieves the best F1 score of 58.5\% and the second best accuracies of 56.7\% and 89.4\%, making it the best choice for CAM-Re2.

%% file: latex/sec6_relatedworks.tex
\section{Related works}
\subsection{Legal article retrieval}
\label{subsec:legal-article-retrieval}
Legal article retrieval
focuses on finding legal articles that are relevant to a specific query. This task has been widely studied, with notable approaches proposed in research such as \cite{louis-spanakis-2022-statutory, paul2022lesicin, louis-etal-2023-finding}. The retrieved legal articles also serve as key components for solving other Legal AI tasks like Legal Question Answering (QA) \cite{louis2024interpretable} and legal judgment prediction \cite{qin2024explicitly}.

To improve the performance of legal article retrieval, some studies have used graph-based methods, connecting legal articles or articles to legal cases. GNNs are often used in these approaches \cite{paul2022lesicin, louis-etal-2023-finding}. However, these methods may not work well for LACD due to two unique challenges outlined in Section~\ref{sec:intro}.

\subsection{LLM generated outputs for legal AI}
LLMs have demonstrated remarkable versatility across a wide range of NLP tasks, leveraging their massive parameterized knowledge \cite{10.5555/3495724.3495883, DBLP:journals/corr/abs-2303-08774, DBLP:journals/corr/abs-2407-21783}.
As a result, many studies have explored using LLM-generated outputs, either as training data \cite{wang-etal-2023-self-instruct} or for making predictions directly during inference \cite{mao-etal-2021-generation, trivedi-etal-2023-interleaving, jiang-etal-2023-active, lee-etal-2024-planrag}.

In the field of Legal AI, some recent studies have shown that incorporating LLM-generated components can improve some Legal AI tasks performance \cite{kim2024self, ma2024leveraging}. 
However, to the best of our knowledge, no prior work has proposed solving LACD by LLM-generated cases.

%% file: latex/sec7_conclusions.tex
\section{Conclusions}
In this paper, 
we proposed a new legal AI task, Legal Article Competition Detection\,(LACD), and construct a dedicate dataset for it.
We propose a novel retriever, CAM-Re2, based on the Case-Augmented Mention Graph\,(CAMGraph) for Korean Law.
We demonstrated that CAM-Re2 is significantly more effective than existing retrievers for the LACD task.


%% file: latex/sec8_limitations.tex
\section{Limitations}
In this paper, we propose CAMGraph as a solution to address the problem of legal competition. However, our approach has several limitations:

First, our methodology has only been validated within the domain of criminal law in Korea. 
Korean criminal law is one of the most extensively studied areas related to legal competition, and it provides a convenient basis for dataset creation. 
However, it is necessary to expand this research to other domains, such as civil, building or administrative law, to address legal competition comprehensively in the future.

Second, CAMGraph only incorporates mention relationships between articles as edges.
For example, methods like G-DSR \cite{louis-etal-2023-finding} utilize tree structures within laws as links, which our approach does not include. 
Whether incorporating such tree structures could effectively solve the LACD problem remains out of scope for this work and requires future investigation.

Lastly, our dataset only contains pairwise competition information, lacking data on competition across a full set of articles. 
To build a more accurate LACD pipeline, developing an LACD benchmark with comprehensive competition information across the entire article set is essential.

%% file: latex/sec9_ethics.tex
\section{Ethical considerations}
\label{sec:ethical-considerations}

Language models have inherent issues with hallucination and the potential to generate biased outputs. In particular, when generating cases for CAMGraph, there is a risk of disproportionately generating cases involving individuals from certain demographic groups, which could lead to harmful biases. 
For instance, CAM-Re2, which utilizes these generated cases, may exhibit a tendency to more effectively retrieve articles (e.g., those related to violence) in lawsuits associated with certain demographic groups. 
To avoid this, only anonymized cases were used as prompts during case generation.
Nevertheless, when developing real-world applications based on our methodology, it is essential to carefully examine the generated cases to identify and mitigate potential biases.

The mention relationships in law was obtained by crawling data from the official website of the Ministry of Government Legislation.
According to Article 7 of the Copyright Act in Korea, legal provisions and compilations of laws created by the government (including link information) are not protected as copyrighted works.

%% file: latex/secA_appendix.tex
\section{Appendix}
\label{sec:appendix}

\subsection{Competitions in the real world}
\label{appendix:competitions-in-the-real-world}

\subsubsection{Hierarchy of laws}
\label{appendix:law-hierarchy}
In Korea, a legal article is included in \textbf{Acts} if and only if the article is enacted by national assembly of Korea. Otherwise, it is classified differently (e.g., enforcement degree, enforcement rule). There exists a hierarchy among Acts, enforcement decrees, and enforcement rules, with Acts being the most authoritative. 
In Korea, if two legal articles of differing hierarchy compete, the lower article must be ignored. 
In this study, we exclusively focus on articles within Acts, and CAMGraph contains 79,615 articles that meet this criterion.

\subsubsection{Solving competitions in Korea}
\label{appendix:solve-competition}

In Korea, if articles $a_1$ and $a_2$ compete with each other, and able to judge some case $c$, one of them is invalidated (i.e., ignored in the judgment).
There are two principles to solve competitions as follows\footnote{\href{https://www.law.go.kr/판례/토석채취허가신청서반려처분무효확인등/(88누6856)}{The supreme court of Korea, 88누6856, 1989. 9. 12.}}: 
\begin{enumerate}
    \item A new law overrides an old law (\textit{lex posterior derogat priori})
    \item A specific law overrides a general law (\textit{lex specialis derogat leges generales})
\end{enumerate}
We explain each principle in Example A.1 and Example A.2, respectively.

\noindent \textit{\textbf{Example A.1. Criminal Act 201 and Narcotics Act 60.}}
As we explained in Section~\ref{sec:intro}, Criminal Act 201 and  Narcotics Act 60 compete with each other, and  thus a crime of using opium is judged by both articles. In terms of time, Criminal Act 201 is relatively old (enacted in 1953) than Narcotics Act 60 (enacted in 2000). Thus, according to principle (1), Narcotics Act 60 overrides Criminal Act 201 (i.e., Criminal Act 201 is ignored in this case).

\noindent \textit{\textbf{Example A.2. Criminal Act 201 and Criminal Act 205.}}
For the case $c_1$ in Example 1, Section~\ref{subsec:definitions}, we can apply not only Criminal Act 205, but also Criminal Act 201 because bob smoked (the same as \textit{used}) opium in his house. Therefore, Criminal Act 201 and 205 are compete with each other and able to judge $c_1$. From the descriptions of each article, Criminal Act 205 judges more general cases than Criminal Act 201 (details are in Example 2, Section 2). Thus, according to principle (2), Criminal Act 201 overrides Criminal Act 201 (i.e., Criminal Act 201 is ignored in this case).

\subsection{Detailed parameters for experiments}
\label{appendix:parameters}

Table \ref{tab:experimental-settings} shows the settings of our experiments.
The reason why the batch size is 16 while the batch size per device is 4, even though we are using an eight GPU machine, is that we are only using 4 GPUs for each experiment.

\begin{table}[htb!]
\renewcommand{\arraystretch}{1.25}
\centering \resizebox{\columnwidth}{!}{
\begin{tabular}{l|c}
\hline
\textbf{Setting} & \textbf{Value} \\ \hline
\textbf{General settings} & \\
Optimizer & Adam \cite{DBLP:journals/corr/KingmaB14}\\
Warmup steps & 500\\
Weight decay & 0 \\
Batch size & 16\\
Batch size per GPUs & 4\\
SEEDs & 0, 42, 2024\\
\hdashline
\textbf{w/ \textit{\textit{enc-cross}}} & \\
learning rate& $5\cdot 10^{-5}$ \\
epochs & 3 \\
\hdashline
\textbf{w/o \textit{\textit{enc-cross}}} & \\
learning rate& $1\cdot 10^{-3}$ \\
epochs & 10 \\
\hline
\textbf{Others} & \\
Context length& 2048 \\
\textit{ProbCal} in Step 3& one layer FFNN \\
\hline
\end{tabular}
}
\caption{Summary of experimental settings. Here, \textit{FFNN} means Feed Forward Neural Networks.}
\label{tab:experimental-settings}
\end{table}

\subsection{Real cases for CAMGraph}
\label{appendix:real-case-for-camgraph}

In this section, we compare two strategies for constructing the nodes in CAMGraph: (1) mixing associated real legal cases with generated cases, and (2) using only generated cases. To achieve this, we collected cases from one of the well-known Korean Legal Benchmark dataset, LBox open~\cite{NEURIPS2022_d15abd14}, and mapped them to corresponding articles. In total, we collect associated cases for 515 articles.

We first build CAMGraph following each strategy, then validated the results by conducting the experiments outlined in Section~\ref{subsec:gnn-only-result}. 
Table~\ref{tab:real-case-experiment} presents the experimental results for both strategies. 
These results demonstrate that mixing real cases significantly underperforms compared to using only generated cases. 
Thus, we construct CAMGraph by following second strategy, using only generated cases. 
We leave investigating the causes of the lower performance and developing methods to effectively utilize real cases for future work.

\begin{table}[htb!]
\centering
\resizebox{\columnwidth}{!}
{
\begin{tabular}{l|cc|cc}
\hline
\multirow{2}{*}{\rule{0pt}{2.5ex}\diagbox{\textbf{Step 3}}{\textbf{Step 1}}}  & \multicolumn{2}{c|}{CAM-Re2 w/ R.C.} & \multicolumn{2}{c}{CAM-Re2 (ours)$^\dagger$} \\
 & F1 & Acc. & F1 & Acc. \\
\hline
CAM-Re2 w/o GNNs& 44.5 & 87.7 & 46.7 & 88.5 \\
CAM-Re2 (ours) & 55.2 & 88.0 & \textbf{64.2} & \textbf{90.4} \\
\hline
\end{tabular}
}
\caption{Step 3 experiment without \textit{enc-cross}. R.C. means real cases. $^\dagger$ means that there are the same results in Section~\ref{subsec:gnn-only-result}.}
\label{tab:real-case-experiment}
\end{table}

\subsection{Multiple cases for enc-bi}
\label{appendix:multiple-cases}

In this section, we compare the effects of employing multiple case augmentations during the training of \textit{enc-bi} by constructing three different encoders: (1) enc-bi in Naïve Re2 (\textbf{N1}), (2) enc-bi in CAM-Re2 with training single generated case (\textbf{single C1}), and (3) enc-bi in CAM-Re2 with training three different generated cases (\textbf{multi C1}). 
To ensure consistency in training computations across all models, N1 and single C1 were trained for three epochs, while multi C1 was trained on three cases ($\times {3}$ training times) within a single epoch ($\times \frac{1}{3}$ training times). 
All other training parameters are the same in Appendix~\ref{appendix:parameters}.

Table~\ref{tab:multiple-case} presents the performance outcomes of these bi-encoders in the Step 3 experiments without GNNs~(see Section~\ref{subsec:gnn-only-result} and Appendix~\ref{appendix:real-case-for-camgraph}).
The results indicate that both single C1 and multi C1 significantly outperform N1, highlighting the impact of Step 1 in the CAM-Re2 method. 
A comparison between single C1 and multi C1 reveals that the latter exhibits a small but positive performance gain in the CAM-Re2 Step 3 results (F1: 60.4\% to 61.9\%; Accuracy: 89.1\% to 89.6\%).

Based on these findings, we conclude that training with multiple cases enhances bi-encoder performance, prompting us to train our bi-encoders on two generated cases. 
A detailed investigation into the impact of training on multiple cases is reserved for future work.

\begin{table}[htb!]
\centering
\resizebox{\columnwidth}{!}
{
\begin{tabular}{l|cc|cc|cc}
\hline
\multirow{2}{*}{\rule{0pt}{2.5ex}\diagbox{\textbf{Step 3}}{\textbf{Step 1}}}  & \multicolumn{2}{c|}{N1} & \multicolumn{2}{c|}{single C1} & \multicolumn{2}{c}{multi C1} \\
 & F1 & Acc. & F1 & Acc. & F1 & Acc. \\
\hline
CAM-Re2 w/o GNNs& 36.1 & 89.4 & 47.5 & 88.7 & 44.4 & \textbf{91.0} \\
CAM-Re2 (ours) & 50.5 & 89.4 & \underline{60.4} & 89.1 & \textbf{61.9} & \underline{89.6} \\
\hline
\end{tabular}
}
\caption{Performance using different \textit{enc-bi} training strategy.}
\label{tab:multiple-case}
\end{table}

\subsection{Acts about crimes}
\label{appendix:criminal-law-acts}

The term \textit{acts about crimes} that we used in Section~\ref{subsec:dataset}, contains following acts. These are selected based on the Korean Bar Exam guidelines\footnote{\href{http://www.lec.co.kr/news/articleView.html?idxno=20966}{Supplementary Acts for Bar Exam, Ministry of Justice, 2011.}}:
\begin{itemize}
    \item Criminal Act
    \item Act on Special Cases Concerning the Punishment of Sexual Crimes
    \item Act on the Aggravated Punishment of Specific Economic Crimes
    \item Act on the Aggravated Punishment of Specific Crimes
    \item Punishment of Violences Act
    \item Act on the Protection of Children and Youth Against Sex Offenses
\end{itemize}

%% file: acl_latex.bbl
\begin{thebibliography}{50}
\expandafter\ifx\csname natexlab\endcsname\relax\def\natexlab#1{#1}\fi

\bibitem[{Araszkiewicz et~al.(2021)Araszkiewicz, Francesconi, and Zurek}]{araszkiewicz2021identification}
Micha{\l} Araszkiewicz, Enrico Francesconi, and Tomasz Zurek. 2021.
\newblock Identification of contradictions in regulation.
\newblock In \emph{Legal Knowledge and Information Systems}, pages 151--160. IOS Press.

\bibitem[{Bai et~al.(2022)Bai, Kadavath, Kundu, Askell, Kernion, Jones, Chen, Goldie, Mirhoseini, McKinnon et~al.}]{bai2022constitutional}
Yuntao Bai, Saurav Kadavath, Sandipan Kundu, Amanda Askell, Jackson Kernion, Andy Jones, Anna Chen, Anna Goldie, Azalia Mirhoseini, Cameron McKinnon, et~al. 2022.
\newblock Constitutional ai: Harmlessness from ai feedback.
\newblock \emph{arXiv preprint arXiv:2212.08073}.

\bibitem[{Bhattacharya et~al.(2020)Bhattacharya, Ghosh, Pal, and Ghosh}]{HierSPCNet}
Paheli Bhattacharya, Kripabandhu Ghosh, Arindam Pal, and Saptarshi Ghosh. 2020.
\newblock \href {https://doi.org/10.1145/3397271.3401191} {Hier-spcnet: A legal statute hierarchy-based heterogeneous network for computing legal case document similarity}.
\newblock In \emph{Proceedings of the 43rd International ACM SIGIR Conference on Research and Development in Information Retrieval}, SIGIR '20, page 1657–1660, New York, NY, USA. Association for Computing Machinery.

\bibitem[{Bommarito~II and Katz(2010)}]{bommarito2010mathematical}
Michael~J Bommarito~II and Daniel~M Katz. 2010.
\newblock A mathematical approach to the study of the united states code.
\newblock \emph{Physica A: Statistical Mechanics and its Applications}, 389(19):4195--4200.

\bibitem[{Brody et~al.(2022)Brody, Alon, and Yahav}]{DBLP:conf/iclr/Brody0Y22}
Shaked Brody, Uri Alon, and Eran Yahav. 2022.
\newblock \href {https://openreview.net/forum?id=F72ximsx7C1} {How attentive are graph attention networks?}
\newblock In \emph{The Tenth International Conference on Learning Representations, {ICLR} 2022, Virtual Event, April 25-29, 2022}. OpenReview.net.

\bibitem[{Brown et~al.(2020)Brown, Mann, Ryder, Subbiah, Kaplan, Dhariwal, Neelakantan, Shyam, Sastry, Askell, Agarwal, Herbert-Voss, Krueger, Henighan, Child, Ramesh, Ziegler, Wu, Winter, Hesse, Chen, Sigler, Litwin, Gray, Chess, Clark, Berner, McCandlish, Radford, Sutskever, and Amodei}]{10.5555/3495724.3495883}
Tom~B. Brown, Benjamin Mann, Nick Ryder, Melanie Subbiah, Jared Kaplan, Prafulla Dhariwal, Arvind Neelakantan, Pranav Shyam, Girish Sastry, Amanda Askell, Sandhini Agarwal, Ariel Herbert-Voss, Gretchen Krueger, Tom Henighan, Rewon Child, Aditya Ramesh, Daniel~M. Ziegler, Jeffrey Wu, Clemens Winter, Christopher Hesse, Mark Chen, Eric Sigler, Mateusz Litwin, Scott Gray, Benjamin Chess, Jack Clark, Christopher Berner, Sam McCandlish, Alec Radford, Ilya Sutskever, and Dario Amodei. 2020.
\newblock Language models are few-shot learners.
\newblock In \emph{Proceedings of the 34th International Conference on Neural Information Processing Systems}, NIPS '20, Red Hook, NY, USA. Curran Associates Inc.

\bibitem[{Coupette et~al.(2021)Coupette, Beckedorf, Hartung, Bommarito, and Katz}]{coupette2021measuring}
Corinna Coupette, Janis Beckedorf, Dirk Hartung, Michael Bommarito, and Daniel~Martin Katz. 2021.
\newblock Measuring law over time: A network analytical framework with an application to statutes and regulations in the united states and germany.
\newblock \emph{Frontiers in Physics}, 9:658463.

\bibitem[{Deng et~al.(2023)Deng, Pei, Kong, Chen, Wei, Li, Ren, Chen, and Ren}]{deng-etal-2023-syllogistic}
Wentao Deng, Jiahuan Pei, Keyi Kong, Zhe Chen, Furu Wei, Yujun Li, Zhaochun Ren, Zhumin Chen, and Pengjie Ren. 2023.
\newblock \href {https://doi.org/10.18653/v1/2023.emnlp-main.864} {Syllogistic reasoning for legal judgment analysis}.
\newblock In \emph{Proceedings of the 2023 Conference on Empirical Methods in Natural Language Processing}, pages 13997--14009, Singapore. Association for Computational Linguistics.

\bibitem[{Dubey et~al.(2024{\natexlab{a}})Dubey, Jauhri, Pandey, Kadian, Al{-}Dahle, Letman, Mathur, Schelten, Yang, Fan, Goyal, Hartshorn, Yang, Mitra, Sravankumar, Korenev, Hinsvark, Rao, Zhang, Rodriguez, Gregerson, Spataru, Rozi{\`{e}}re, Biron, Tang, Chern, Caucheteux, Nayak, Bi, Marra, McConnell, Keller, Touret, Wu, Wong, Ferrer, Nikolaidis, Allonsius, Song, Pintz, Livshits, Esiobu, Choudhary, Mahajan, Garcia{-}Olano, Perino, Hupkes, Lakomkin, AlBadawy, Lobanova, Dinan, Smith, Radenovic, Zhang, Synnaeve, Lee, Anderson, Nail, Mialon, Pang, Cucurell, Nguyen, Korevaar, Xu, Touvron, Zarov, Ibarra, Kloumann, Misra, Evtimov, Copet, Lee, Geffert, Vranes, Park, Mahadeokar, Shah, van~der Linde, Billock, Hong, Lee, Fu, Chi, Huang, Liu, Wang, Yu, Bitton, Spisak, Park, Rocca, Johnstun, Saxe, Jia, Alwala, Upasani, Plawiak, Li, Heafield, Stone, and et~al.}]{DBLP:journals/corr/abs-2407-21783}
Abhimanyu Dubey, Abhinav Jauhri, Abhinav Pandey, Abhishek Kadian, Ahmad Al{-}Dahle, Aiesha Letman, Akhil Mathur, Alan Schelten, Amy Yang, Angela Fan, Anirudh Goyal, Anthony Hartshorn, Aobo Yang, Archi Mitra, Archie Sravankumar, Artem Korenev, Arthur Hinsvark, Arun Rao, Aston Zhang, Aur{\'{e}}lien Rodriguez, Austen Gregerson, Ava Spataru, Baptiste Rozi{\`{e}}re, Bethany Biron, Binh Tang, Bobbie Chern, Charlotte Caucheteux, Chaya Nayak, Chloe Bi, Chris Marra, Chris McConnell, Christian Keller, Christophe Touret, Chunyang Wu, Corinne Wong, Cristian~Canton Ferrer, Cyrus Nikolaidis, Damien Allonsius, Daniel Song, Danielle Pintz, Danny Livshits, David Esiobu, Dhruv Choudhary, Dhruv Mahajan, Diego Garcia{-}Olano, Diego Perino, Dieuwke Hupkes, Egor Lakomkin, Ehab AlBadawy, Elina Lobanova, Emily Dinan, Eric~Michael Smith, Filip Radenovic, Frank Zhang, Gabriel Synnaeve, Gabrielle Lee, Georgia~Lewis Anderson, Graeme Nail, Gr{\'{e}}goire Mialon, Guan Pang, Guillem Cucurell, Hailey Nguyen, Hannah Korevaar, Hu~Xu, Hugo
  Touvron, Iliyan Zarov, Imanol~Arrieta Ibarra, Isabel~M. Kloumann, Ishan Misra, Ivan Evtimov, Jade Copet, Jaewon Lee, Jan Geffert, Jana Vranes, Jason Park, Jay Mahadeokar, Jeet Shah, Jelmer van~der Linde, Jennifer Billock, Jenny Hong, Jenya Lee, Jeremy Fu, Jianfeng Chi, Jianyu Huang, Jiawen Liu, Jie Wang, Jiecao Yu, Joanna Bitton, Joe Spisak, Jongsoo Park, Joseph Rocca, Joshua Johnstun, Joshua Saxe, Junteng Jia, Kalyan~Vasuden Alwala, Kartikeya Upasani, Kate Plawiak, Ke~Li, Kenneth Heafield, Kevin Stone, and et~al. 2024{\natexlab{a}}.
\newblock \href {https://doi.org/10.48550/ARXIV.2407.21783} {The llama 3 herd of models}.
\newblock \emph{CoRR}, abs/2407.21783.

\bibitem[{Dubey et~al.(2024{\natexlab{b}})Dubey, Jauhri, Pandey, Kadian, Al-Dahle, Letman, Mathur, Schelten, Yang, Fan et~al.}]{dubey2024llama}
Abhimanyu Dubey, Abhinav Jauhri, Abhinav Pandey, Abhishek Kadian, Ahmad Al-Dahle, Aiesha Letman, Akhil Mathur, Alan Schelten, Amy Yang, Angela Fan, et~al. 2024{\natexlab{b}}.
\newblock The llama 3 herd of models.
\newblock \emph{arXiv preprint arXiv:2407.21783}.

\bibitem[{Feng et~al.(2022{\natexlab{a}})Feng, Li, and Ng}]{feng2022legal}
Yi~Feng, Chuanyi Li, and Vincent Ng. 2022{\natexlab{a}}.
\newblock Legal judgment prediction: A survey of the state of the art.
\newblock In \emph{IJCAI}, pages 5461--5469.

\bibitem[{Feng et~al.(2022{\natexlab{b}})Feng, Li, and Ng}]{feng-etal-2022-legal}
Yi~Feng, Chuanyi Li, and Vincent Ng. 2022{\natexlab{b}}.
\newblock \href {https://doi.org/10.18653/v1/2022.acl-long.48} {Legal judgment prediction via event extraction with constraints}.
\newblock In \emph{Proceedings of the 60th Annual Meeting of the Association for Computational Linguistics (Volume 1: Long Papers)}, pages 648--664, Dublin, Ireland. Association for Computational Linguistics.

\bibitem[{Glass et~al.(2022)Glass, Rossiello, Chowdhury, Naik, Cai, and Gliozzo}]{glass-etal-2022-re2g}
Michael Glass, Gaetano Rossiello, Md~Faisal~Mahbub Chowdhury, Ankita Naik, Pengshan Cai, and Alfio Gliozzo. 2022.
\newblock \href {https://doi.org/10.18653/v1/2022.naacl-main.194} {{R}e2{G}: Retrieve, rerank, generate}.
\newblock In \emph{Proceedings of the 2022 Conference of the North American Chapter of the Association for Computational Linguistics: Human Language Technologies}, pages 2701--2715, Seattle, United States. Association for Computational Linguistics.

\bibitem[{Hamilton et~al.(2017)Hamilton, Ying, and Leskovec}]{DBLP:conf/nips/HamiltonYL17}
William~L. Hamilton, Zhitao Ying, and Jure Leskovec. 2017.
\newblock \href {https://proceedings.neurips.cc/paper/2017/hash/5dd9db5e033da9c6fb5ba83c7a7ebea9-Abstract.html} {Inductive representation learning on large graphs}.
\newblock In \emph{Advances in Neural Information Processing Systems 30: Annual Conference on Neural Information Processing Systems 2017, December 4-9, 2017, Long Beach, CA, {USA}}, pages 1024--1034.

\bibitem[{Holzenberger et~al.(2020)Holzenberger, Blair-Stanek, and Durme}]{Holzenberger2020ADF}
Nils Holzenberger, Andrew Blair-Stanek, and Benjamin~Van Durme. 2020.
\newblock \href {https://api.semanticscholar.org/CorpusID:218581117} {A dataset for statutory reasoning in tax law entailment and question answering}.
\newblock In \emph{NLLP@KDD}.

\bibitem[{Hua et~al.(2024)Hua, Yang, Jin, Li, Cheng, Tang, and Zhang}]{hua2024trustagent}
Wenyue Hua, Xianjun Yang, Mingyu Jin, Zelong Li, Wei Cheng, Ruixiang Tang, and Yongfeng Zhang. 2024.
\newblock Trustagent: Towards safe and trustworthy llm-based agents.
\newblock In \emph{Findings of the Association for Computational Linguistics: EMNLP 2024}, pages 10000--10016.

\bibitem[{Hwang et~al.(2022)Hwang, Lee, Cho, Lee, and Seo}]{NEURIPS2022_d15abd14}
Wonseok Hwang, Dongjun Lee, Kyoungyeon Cho, Hanuhl Lee, and Minjoon Seo. 2022.
\newblock \href {https://proceedings.neurips.cc/paper_files/paper/2022/file/d15abd14d5894eebd185b756541d420e-Paper-Datasets_and_Benchmarks.pdf} {A multi-task benchmark for korean legal language understanding and judgement prediction}.
\newblock In \emph{Advances in Neural Information Processing Systems}, volume~35, pages 32537--32551. Curran Associates, Inc.

\bibitem[{Jiang et~al.(2023)Jiang, Xu, Gao, Sun, Liu, Dwivedi-Yu, Yang, Callan, and Neubig}]{jiang-etal-2023-active}
Zhengbao Jiang, Frank Xu, Luyu Gao, Zhiqing Sun, Qian Liu, Jane Dwivedi-Yu, Yiming Yang, Jamie Callan, and Graham Neubig. 2023.
\newblock \href {https://doi.org/10.18653/v1/2023.emnlp-main.495} {Active retrieval augmented generation}.
\newblock In \emph{Proceedings of the 2023 Conference on Empirical Methods in Natural Language Processing}, pages 7969--7992, Singapore. Association for Computational Linguistics.

\bibitem[{Karpukhin et~al.(2020)Karpukhin, Oguz, Min, Lewis, Wu, Edunov, Chen, and Yih}]{karpukhin-etal-2020-dense}
Vladimir Karpukhin, Barlas Oguz, Sewon Min, Patrick Lewis, Ledell Wu, Sergey Edunov, Danqi Chen, and Wen-tau Yih. 2020.
\newblock \href {https://doi.org/10.18653/v1/2020.emnlp-main.550} {Dense passage retrieval for open-domain question answering}.
\newblock In \emph{Proceedings of the 2020 Conference on Empirical Methods in Natural Language Processing (EMNLP)}, pages 6769--6781, Online. Association for Computational Linguistics.

\bibitem[{Katz et~al.(2020)Katz, Coupette, Beckedorf, and Hartung}]{katz2020complex}
Daniel~Martin Katz, Corinna Coupette, Janis Beckedorf, and Dirk Hartung. 2020.
\newblock Complex societies and the growth of the law.
\newblock \emph{Scientific reports}, 10(1):18737.

\bibitem[{Kim et~al.(2024)Kim, Jung, and Koo}]{kim2024self}
Minju Kim, Haein Jung, and Myoung-Wan Koo. 2024.
\newblock Self-expertise: Knowledge-based instruction dataset augmentation for a legal expert language model.
\newblock In \emph{Findings of the Association for Computational Linguistics: NAACL 2024}, pages 1098--1112.

\bibitem[{Kim(2005)}]{kim2005Fallgruppen}
Seong-Don Kim. 2005.
\newblock Fallgruppen der gesetzeskonkurrenz und ihre bewertungsmethode.
\newblock \emph{Korean Lawyers Association Journal}, 54(1):29--67.

\bibitem[{Kingma and Ba(2015)}]{DBLP:journals/corr/KingmaB14}
Diederik~P. Kingma and Jimmy Ba. 2015.
\newblock \href {http://arxiv.org/abs/1412.6980} {Adam: {A} method for stochastic optimization}.
\newblock In \emph{3rd International Conference on Learning Representations, {ICLR} 2015, San Diego, CA, USA, May 7-9, 2015, Conference Track Proceedings}.

\bibitem[{Kipf and Welling(2017)}]{DBLP:conf/iclr/KipfW17}
Thomas~N. Kipf and Max Welling. 2017.
\newblock \href {https://openreview.net/forum?id=SJU4ayYgl} {Semi-supervised classification with graph convolutional networks}.
\newblock In \emph{5th International Conference on Learning Representations, {ICLR} 2017, Toulon, France, April 24-26, 2017, Conference Track Proceedings}. OpenReview.net.

\bibitem[{Lee et~al.(2024)Lee, An, and Kim}]{lee-etal-2024-planrag}
Myeonghwa Lee, Seonho An, and Min-Soo Kim. 2024.
\newblock \href {https://doi.org/10.18653/v1/2024.naacl-long.364} {{P}lan{RAG}: A plan-then-retrieval augmented generation for generative large language models as decision makers}.
\newblock In \emph{Proceedings of the 2024 Conference of the North American Chapter of the Association for Computational Linguistics: Human Language Technologies (Volume 1: Long Papers)}, pages 6537--6555, Mexico City, Mexico. Association for Computational Linguistics.

\bibitem[{Liu et~al.(2023)Liu, Wu, Zhang, Sun, Lu, Wu, and Kuang}]{liu2023ml}
Yifei Liu, Yiquan Wu, Yating Zhang, Changlong Sun, Weiming Lu, Fei Wu, and Kun Kuang. 2023.
\newblock Ml-ljp: multi-law aware legal judgment prediction.
\newblock In \emph{Proceedings of the 46th international ACM SIGIR conference on research and development in information retrieval}, pages 1023--1034.

\bibitem[{Louis and Spanakis(2022)}]{louis-spanakis-2022-statutory}
Antoine Louis and Gerasimos Spanakis. 2022.
\newblock \href {https://doi.org/10.18653/v1/2022.acl-long.468} {A statutory article retrieval dataset in {F}rench}.
\newblock In \emph{Proceedings of the 60th Annual Meeting of the Association for Computational Linguistics (Volume 1: Long Papers)}, pages 6789--6803, Dublin, Ireland. Association for Computational Linguistics.

\bibitem[{Louis et~al.(2023)Louis, van Dijck, and Spanakis}]{louis-etal-2023-finding}
Antoine Louis, Gijs van Dijck, and Gerasimos Spanakis. 2023.
\newblock \href {https://doi.org/10.18653/v1/2023.eacl-main.203} {Finding the law: Enhancing statutory article retrieval via graph neural networks}.
\newblock In \emph{Proceedings of the 17th Conference of the European Chapter of the Association for Computational Linguistics}, pages 2761--2776, Dubrovnik, Croatia. Association for Computational Linguistics.

\bibitem[{Louis et~al.(2024)Louis, van Dijck, and Spanakis}]{louis2024interpretable}
Antoine Louis, Gijs van Dijck, and Gerasimos Spanakis. 2024.
\newblock Interpretable long-form legal question answering with retrieval-augmented large language models.
\newblock In \emph{Proceedings of the AAAI Conference on Artificial Intelligence}, volume~38, pages 22266--22275.

\bibitem[{Ma et~al.(2024)Ma, Chen, Chu, and Mao}]{ma2024leveraging}
Shengjie Ma, Chong Chen, Qi~Chu, and Jiaxin Mao. 2024.
\newblock Leveraging large language models for relevance judgments in legal case retrieval.
\newblock \emph{arXiv preprint arXiv:2403.18405}.

\bibitem[{Mao et~al.(2021)Mao, He, Liu, Shen, Gao, Han, and Chen}]{mao-etal-2021-generation}
Yuning Mao, Pengcheng He, Xiaodong Liu, Yelong Shen, Jianfeng Gao, Jiawei Han, and Weizhu Chen. 2021.
\newblock \href {https://doi.org/10.18653/v1/2021.acl-long.316} {Generation-augmented retrieval for open-domain question answering}.
\newblock In \emph{Proceedings of the 59th Annual Meeting of the Association for Computational Linguistics and the 11th International Joint Conference on Natural Language Processing (Volume 1: Long Papers)}, pages 4089--4100, Online. Association for Computational Linguistics.

\bibitem[{{Ministry of Government Legislation}(2023)}]{moleg2023standards}
{Ministry of Government Legislation}. 2023.
\newblock \href {https://www.moleg.go.kr/menu.es?mid=a10105030000} {Standards for legislative drafting and review}.
\newblock Accessed: 2024-12-15.

\bibitem[{Nogueira and Cho(2019)}]{nogueira2019passage}
Rodrigo Nogueira and Kyunghyun Cho. 2019.
\newblock Passage re-ranking with bert.
\newblock \emph{arXiv preprint arXiv:1901.04085}.

\bibitem[{OpenAI(2023)}]{DBLP:journals/corr/abs-2303-08774}
OpenAI. 2023.
\newblock \href {https://doi.org/10.48550/ARXIV.2303.08774} {{GPT-4} technical report}.
\newblock \emph{CoRR}, abs/2303.08774.

\bibitem[{Park and Kim(2021)}]{jangwon_park_2021_5654154}
Jangwon Park and Donggyu Kim. 2021.
\newblock \href {https://doi.org/10.5281/zenodo.5654154} {Kobigbird: Pretrained bigbird model for korean}.

\bibitem[{Paul et~al.(2024)Paul, Bhatt, Goyal, and Ghosh}]{10.1145/3626772.3657879}
Shounak Paul, Rajas Bhatt, Pawan Goyal, and Saptarshi Ghosh. 2024.
\newblock \href {https://doi.org/10.1145/3626772.3657879} {Legal statute identification: A case study using state-of-the-art datasets and methods}.
\newblock In \emph{Proceedings of the 47th International ACM SIGIR Conference on Research and Development in Information Retrieval}, SIGIR '24, page 2231–2240, New York, NY, USA. Association for Computing Machinery.

\bibitem[{Paul et~al.(2022)Paul, Goyal, and Ghosh}]{paul2022lesicin}
Shounak Paul, Pawan Goyal, and Saptarshi Ghosh. 2022.
\newblock Lesicin: A heterogeneous graph-based approach for automatic legal statute identification from indian legal documents.
\newblock In \emph{Proceedings of the AAAI conference on artificial intelligence}, volume~36, pages 11139--11146.

\bibitem[{Qin et~al.(2024)Qin, Cao, Yu, Si, Chen, and Xu}]{qin2024explicitly}
Weicong Qin, Zelin Cao, Weijie Yu, Zihua Si, Sirui Chen, and Jun Xu. 2024.
\newblock \href {https://doi.org/10.1145/3626772.3657717} {Explicitly integrating judgment prediction with legal document retrieval: A law-guided generative approach}.
\newblock In \emph{Proceedings of the 47th International ACM SIGIR Conference on Research and Development in Information Retrieval}, SIGIR '24, page 2210–2220, New York, NY, USA. Association for Computing Machinery.

\bibitem[{Robertson et~al.(2009)Robertson, Zaragoza et~al.}]{robertson2009probabilistic}
Stephen Robertson, Hugo Zaragoza, et~al. 2009.
\newblock The probabilistic relevance framework: Bm25 and beyond.
\newblock \emph{Foundations and Trends{\textregistered} in Information Retrieval}, 3(4):333--389.

\bibitem[{Shao et~al.(2020)Shao, Mao, Liu, Ma, Satoh, Zhang, and Ma}]{shao2020bert}
Yunqiu Shao, Jiaxin Mao, Yiqun Liu, Weizhi Ma, Ken Satoh, Min Zhang, and Shaoping Ma. 2020.
\newblock Bert-pli: Modeling paragraph-level interactions for legal case retrieval.
\newblock In \emph{IJCAI}, pages 3501--3507.

\bibitem[{Song et~al.(2024)Song, Kim, Lee, Kim, and Thorne}]{song-etal-2024-re3val}
EuiYul Song, Sangryul Kim, Haeju Lee, Joonkee Kim, and James Thorne. 2024.
\newblock \href {https://aclanthology.org/2024.findings-eacl.27} {Re3val: Reinforced and reranked generative retrieval}.
\newblock In \emph{Findings of the Association for Computational Linguistics: EACL 2024}, pages 393--409, St. Julian{'}s, Malta. Association for Computational Linguistics.

\bibitem[{Sun et~al.(2023)Sun, Xu, Zhang, Dong, and Wen}]{sun2023law}
Zhongxiang Sun, Jun Xu, Xiao Zhang, Zhenhua Dong, and Ji-Rong Wen. 2023.
\newblock Law article-enhanced legal case matching: A causal learning approach.
\newblock In \emph{Proceedings of the 46th International ACM SIGIR Conference on Research and Development in Information Retrieval}, pages 1549--1558.

\bibitem[{Trivedi et~al.(2023)Trivedi, Balasubramanian, Khot, and Sabharwal}]{trivedi-etal-2023-interleaving}
Harsh Trivedi, Niranjan Balasubramanian, Tushar Khot, and Ashish Sabharwal. 2023.
\newblock \href {https://doi.org/10.18653/v1/2023.acl-long.557} {Interleaving retrieval with chain-of-thought reasoning for knowledge-intensive multi-step questions}.
\newblock In \emph{Proceedings of the 61st Annual Meeting of the Association for Computational Linguistics (Volume 1: Long Papers)}, pages 10014--10037, Toronto, Canada. Association for Computational Linguistics.

\bibitem[{Wang et~al.(2023)Wang, Kordi, Mishra, Liu, Smith, Khashabi, and Hajishirzi}]{wang-etal-2023-self-instruct}
Yizhong Wang, Yeganeh Kordi, Swaroop Mishra, Alisa Liu, Noah~A. Smith, Daniel Khashabi, and Hannaneh Hajishirzi. 2023.
\newblock \href {https://doi.org/10.18653/v1/2023.acl-long.754} {Self-instruct: Aligning language models with self-generated instructions}.
\newblock In \emph{Proceedings of the 61st Annual Meeting of the Association for Computational Linguistics (Volume 1: Long Papers)}, pages 13484--13508, Toronto, Canada. Association for Computational Linguistics.

\bibitem[{Wu et~al.(2020)Wu, Petroni, Josifoski, Riedel, and Zettlemoyer}]{wu-etal-2020-scalable}
Ledell Wu, Fabio Petroni, Martin Josifoski, Sebastian Riedel, and Luke Zettlemoyer. 2020.
\newblock \href {https://doi.org/10.18653/v1/2020.emnlp-main.519} {Scalable zero-shot entity linking with dense entity retrieval}.
\newblock In \emph{Proceedings of the 2020 Conference on Empirical Methods in Natural Language Processing (EMNLP)}, pages 6397--6407, Online. Association for Computational Linguistics.

\bibitem[{Xu et~al.(2020)Xu, Wang, Chen, Pan, Wang, and Zhao}]{xu-etal-2020-distinguish}
Nuo Xu, Pinghui Wang, Long Chen, Li~Pan, Xiaoyan Wang, and Junzhou Zhao. 2020.
\newblock \href {https://doi.org/10.18653/v1/2020.acl-main.280} {Distinguish confusing law articles for legal judgment prediction}.
\newblock In \emph{Proceedings of the 58th Annual Meeting of the Association for Computational Linguistics}, pages 3086--3095, Online. Association for Computational Linguistics.

\bibitem[{Yang et~al.(2024)Yang, Yang, Hui, Zheng, Yu, Zhou, Li, Li, Liu, Huang et~al.}]{yang2024qwen2}
An~Yang, Baosong Yang, Binyuan Hui, Bo~Zheng, Bowen Yu, Chang Zhou, Chengpeng Li, Chengyuan Li, Dayiheng Liu, Fei Huang, et~al. 2024.
\newblock Qwen2 technical report.
\newblock \emph{arXiv preprint arXiv:2407.10671}.

\bibitem[{Yoon(2005)}]{Yoon2005Grund}
Dong-Ho Yoon. 2005.
\newblock Grundforschung zur reform der sonderstrafgesetzbuche.
\newblock \emph{Korean Institute of Criminology and Justice}, pages 9--282.

\bibitem[{Yue et~al.(2021)Yue, Liu, Jin, Wu, Zhang, An, Cheng, Yin, and Wu}]{10.1145/3404835.3462826}
Linan Yue, Qi~Liu, Binbin Jin, Han Wu, Kai Zhang, Yanqing An, Mingyue Cheng, Biao Yin, and Dayong Wu. 2021.
\newblock \href {https://doi.org/10.1145/3404835.3462826} {Neurjudge: A circumstance-aware neural framework for legal judgment prediction}.
\newblock In \emph{Proceedings of the 44th International ACM SIGIR Conference on Research and Development in Information Retrieval}, SIGIR '21, page 973–982, New York, NY, USA. Association for Computing Machinery.

\bibitem[{Zhu et~al.(2023)Zhu, Yu, Jin, Hou, Li, and Sui}]{zhu-etal-2023-learn}
Fangwei Zhu, Jifan Yu, Hailong Jin, Lei Hou, Juanzi Li, and Zhifang Sui. 2023.
\newblock \href {https://doi.org/10.18653/v1/2023.findings-acl.690} {Learn to not link: Exploring {NIL} prediction in entity linking}.
\newblock In \emph{Findings of the Association for Computational Linguistics: ACL 2023}, pages 10846--10860, Toronto, Canada. Association for Computational Linguistics.

\end{thebibliography}
